\documentclass{article}
\usepackage[utf8]{inputenc}
\usepackage[T1]{fontenc}
\usepackage{spconf,amsmath,graphicx}

\title{Jigsaw Puzzle Solving Using Local Feature Co-Occurrences\\in Deep Neural Networks}
\name{Marie-Morgane Paumard$^1$, David Picard$^{1,2}$, Hedi Tabia$^1$ \thanks{This work is supported by the Fondation des sciences du patrimoine, LabEx PATRIMA ANR-10-LABX-0094-01}}
\address{$^1$ETIS, UMR 8051, Université Paris Seine, Université Cergy-Pontoise, ENSEA, CNRS\\
$^2$Sorbonne Université, CNRS, Laboratoire d'Informatique de Paris 6, LIP6, F-75005 Paris
}

\begin{document}
%
\maketitle
\begin{abstract}
Archaeologists are in dire need of automated object reconstruction methods. Fragments reassembly is close to puzzle problems, which may be solved by computer vision algorithms. As they are often beaten on most image related tasks by deep learning algorithms, we study a classification method that can solve jigsaw puzzles.
In this paper, we focus on classifying the relative position: given a couple of fragments, we compute their local relation (e.g. on top). We propose several enhancements over the state of the art in this domain, which is outperformed by our method by 25\%. We propose an original dataset composed of pictures from the Metropolitan Museum of Art. We propose a greedy reconstruction method based on the predicted relative positions.
\end{abstract}
\section{Introduction}
\label{sec:intro}
The reconstruction of pieces of art from shards is a time-consuming task for archaeologists. Close to puzzle solving problems, it may be automated. On the one hand, the computer vision algorithms struggle with those tasks. The dataset has to be scrupulously annotated by experts to reach a plausible solution, especially when the object fragments are lost, degraded or mixed among non-relevant fragments. Even so, the false-positive rate is still significant. On the other hand, deep learning algorithms are seen as an promising alternative, as they surpass other methods in most image-related tasks.

\begin{figure}[htb]
  \centering
  \centerline{\includegraphics[width=8.5cm]{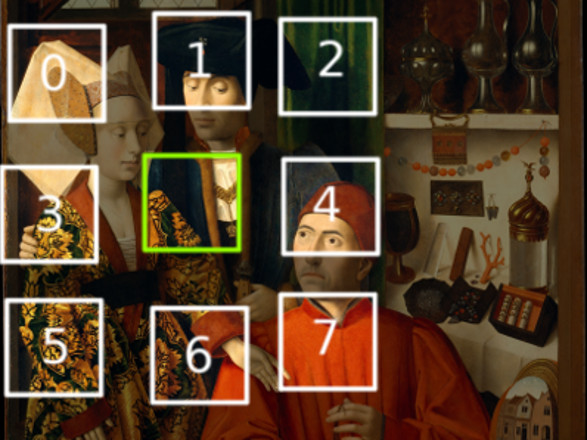}}
  \caption{Example of fragments extracted using the randomized grid pattern on the MET dataset. Labels are the classes of the relative position with respect to the central fragment.}
  \label{fig:demo}
\end{figure}

In this paper, we consider the image reassembly, which can be seen as solving a jigsaw puzzle: given two image fragments, we want to predict the relative position of the second fragment with respect to the first one. To solve this task, we investigate the setup proposed by Doersch et al. \cite{doersch}. Given an image, we extract same-size squared 2D-tiles in a randomized grid pattern (see Figure \ref{fig:demo}). Visual features are then extracted from both fragments using a Convolutional Neural Network (CNN). These features are the combined and fed to a classifier in order to predict the correct relative position.

However, the authors of \cite{doersch} are not interested in solving the jigsaw puzzle problem \textit{per se}, but merely in using it as a pretraining of generic visual features.

In order to solve the jigsaw puzzle, we extend the work of Doersch et al. \cite{doersch}. Our contributions are the following: First, we propose a simpler, yet more effective, CNN architecture to extract visual features. Then, we propose a new combination scheme based on the Kronecker product that is able to better take into account correlations between localized parts of the fragments. We also propose a new dataset more closely related to cultural heritage puzzle solving tasks, which consists of 14,000 images from the Metropolitan Museum of Art (MET) archives. With our contributions, we obtain state of the art results on both the original ImageNet dataset proposed in \cite{doersch} and our new MET dataset.

The remaining of this paper is organized as follows: in the next section, we present related work and contextual information on puzzle solving and fragment reassembly. Then, we detail our method in Section \ref{sec:mth}. Finally, we describe our new dataset and we examine our experimental results in Section \ref{sec:res}.

\begin{figure*}[htb]
  \centering
  \centerline{\includegraphics[width=\textwidth]{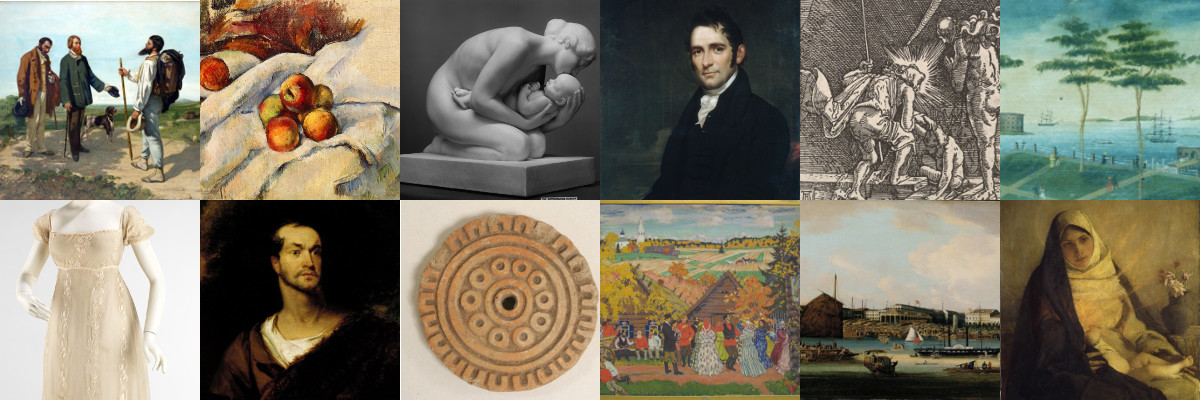}}
  \caption{Examples of images from the MET dataset.}
  \label{fig:met}
\end{figure*}

\section{Related work}
\label{sec:soa}

Whether it comes to precisely align fragments or approximate a relative position, archaeological object reconstruction attracts numerous researchers, as shown by Rasheed and Nordin in \cite{survey,rasheed}. When most reassembly work is based on semi-automated methods, some stands out by the use of automated reconstruction, such as \cite{mcbride,jampy,zhu,glassnegative}. These are mainly inspired by the research on puzzle solving, especially on jigsaw puzzles \cite{jigsaw1,jigsaw2,jigsaw3,gur}. Those methods study missing fragments or various-sized tiles, belonging to annotated datasets. They perform well on small datasets but poorly when the fragments are mixed from similar sources. Moreover, these methods are often very slow.

Since 2015, the deep learning community uses puzzle solving as a pretext task, proposing a reasonable alternative to data labeling. In \cite{doersch}, Doersch et al. introduce puzzle solving as a pretraining task for objects classification and detection. Their algorithm outperforms other unsupervised methods, which illustrates that CNN are able to learn object representations from the relative positions of the image part. Based on \cite{doersch}, Noroozi and Favaro \cite{noroozi} propose a network that observes all the nine tiles arranged in a grid pattern to obtain a precise object representation. They claim that the ambiguities may be wiped out more effectively when all fragments are examined. However, this leads to a much more complex classification due to the huge number of fragment orderings.

In our case, as we are not interested in building generic image features, but in solving approximatively the jigsaw puzzle itself. As such, the setup of \cite{noroozi} is impractical as it requires all fragments in order to make a prediction. This is unrealistic in cultural heritage where missing fragments are very common. However, the correlations between localized parts of the fragments (e.g., the correlation between the right part of the central fragment and the left part of the right fragment) are not taken into account in \cite{doersch,noroozi}, whereas we argue this information is essential to successful classification. This is what we investigate in our method.

\section{Proposed Method}
\label{sec:mth}

In this section, we detail our proposed method. We start by presenting the Feature Extraction Network that is shared between the two fragments. Then, we describe our propositions to combine the features of both fragments leading to the classification stage. Finally, we present a greedy algorithm to solve the jigsaw puzzle given any number of fragments.

\subsection{Feature Extraction Network architecture}
We use a CNN to compute the features associated with the fragments. Each fragment of size $96 \times 96$ pixel is processed such that pixels values vary between $-1$ and 1. Our architecture takes inspiration from VGG \cite{vgg} and is composed of a sequence of $3 \times 3$ convolutions followed by batch-normalizations \cite{bn}, ReLU activations \cite{relu} and max-poolings. The full architecture is shown on Table \ref{tab:archi}.

\begin{table}[htb]
  \centering
  \begin{tabular}{|c|c|c|}
    \hline
    Layer & Shape & \# parameters \\
    \hline
    Input & $96 \times 96 \times 3$ & 0 \\
    Conv+BN+ReLU & $96 \times 96 \times 32$ & 1k \\
    Maxpooling & $48 \times 48 \times 32$ & - \\
    Conv+BN+ReLU & $48 \times 48 \times 64$ & 19k \\
    Maxpooling & $24 \times 24 \times 64$ & -\\
    Conv+BN+ReLU & $24 \times 24 \times 128$ & 74k \\
    Maxpooling & $12 \times 12 \times 128$ & - \\
    Conv+BN+ReLU & $12 \times 12 \times 256$ & 296k \\
    Maxpooling & $6 \times 6 \times 256$ & -\\
    Conv+BN+ReLU & $6 \times 6 \times 512$ & 1.2M \\
    Maxpooling & $3 \times 3 \times 512$ & -\\
    Fully Connected+BN & 512 &  2.4M \\
    \hline
  \end{tabular}
  \caption{Architecture of the Feature Extraction Network. Conv: 3$\times$3 convolution, BN: Batch-Normalization, ReLU: ReLU activation.}
  \label{tab:archi}
\end{table}

Our Feature Extraction Network ends with a fully connected layer that allows preserving the spatial layout of the input fragment. Although it is more costly than the worse popular global pooling layers \cite{global}, we conjecture that keeping the layout is essential to successfully predict the relative position of fragments.

\subsection{Combination Layer}
\begin{figure}[htb]
s  \centerline{\includegraphics[width=8.5cm]{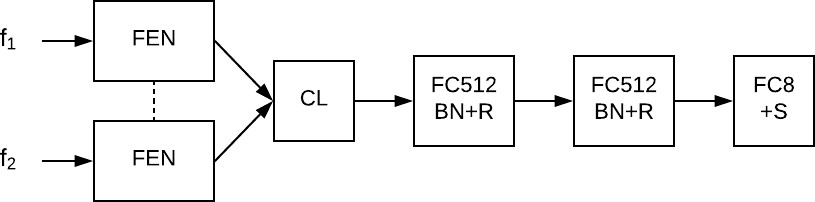}}
  \caption{Full network architecture. FEN: Feature Extractor Network. CL Combination Layer. FC: Fully Connected. BN: Batch-Normalization. R: ReLU activation. S: Softmax activation.}
  \label{fig:fna}
\end{figure}

To predict the relative position of a fragment with respect to another one, we extract features for both fragments using the same feature extraction network. These two features are then combined using a combination layer and processed by a neural network as shown in Figure \ref{fig:fna}.

In \cite{doersch}, the authors propose to concatenate both features in the combination layer. This leads the subsequent fully connected layer to perform a linear combination of the fragment features:
\[\forall i, y_i=\sum_m \alpha_{i,m} \phi_m(f_1) + \sum_m \beta_{i,m} \phi_m(f_2)\]
where $y_i$ is the output of the neuron $i$ in the fully connected layer, $\phi(f_1)$ is the feature of the first fragment (respectively, $\phi(f_2)$ for the second fragment) and $\alpha_{i,m}, \beta_{i,m}$ are the weights of neuron $i$.

Such a linear combination does not highlight co-occurrences of features, although it can be argued that subsequent layers with non-linearities may eventually be able to achieve similar results.

To circumvent these problems, we propose to use the Kronecker product in the combination layer. The output of the fully connected layer is then:
\[\forall i, y_i=\sum_{m,n}\alpha_{i,m,n}\phi_m(f_1)\phi_n(f_2),\]
with $\alpha_{i,m,n}$ being the weights of neuron $i$. The Kronecker product enables to explicitly model the co-occurrences between features of both fragments. This comes however at the cost of an increased number of parameters.

The output of the full network consists of a fully connected layer with $k$ neurons followed by a Softmax activation, corresponding to the probabilities of the $k$ possible relative locations.
The full network (the Feature Extraction part and the Classification part) is trained at once using stochastic gradient descent on batches of fragments pairs.

\subsection{Puzzle solving}
In order to solve the puzzle problem, we consider the case where given a central fragment, we want to assign each of the remaining 8 fragments to its correct location. For each fragment, we compute the probabilities of assignment using the full network. This results in an  $8\times 8$ matrix where each row is a fragment and each column is a possible location. Solving the puzzle problem corresponds then to an assignment problem where we have to pick 8 values from the matrix (only one per row/column) such that their sum is maximized. We propose a greedy algorithm where we iteratively pick the maximum value and remove its corresponding row and column. Remark that this problem corresponds to a graph-cut problem for which much more involved algorithms exist. However, we found out that our greedy algorithm provides correct results in our case.

\section{Experiments}
\label{sec:res}

\begin{figure*}[thb]
  \centering
  \centerline{\includegraphics[width=\textwidth]{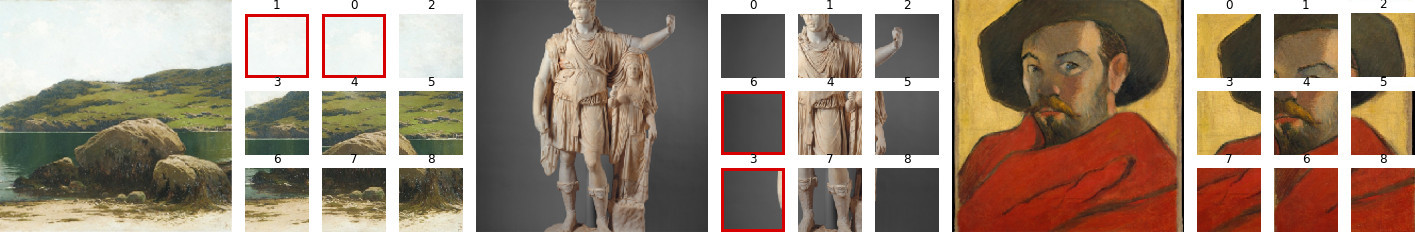}}
  \caption{Examples of reconstructions using the greedy algorithm on images taken from the MET dataset. The red outlined fragments are misplaced.}
  \label{fig:puzzle}
\end{figure*}

In this section, we first describe our new dataset and experimental setup. Then, we comment on the results comparing our approach to \cite{doersch}. Finally, we give some qualitative results on puzzle solving using the greedy algorithm.

\subsection{MET dataset and experimental setup}

In order to be closer to our aimed application regarding cultural heritage puzzle solving, we propose a new dataset consisting of images from the Metropolitan Museum of Art. We collect 14,000 open-source images of paintings and pieces of art. Contrarily to ImageNet, the quality of the sensors taking the pictures allows avoiding the unbalanced colors distribution. Images from the MET dataset are shown in Figure \ref{fig:met}.

We use 10k images for training and evaluate the performances on the remaining 4k.
During training, each image from the training set is resized and square-cropped so that its size is $398 \times 398$ pixels. We divide it into 9 parts separated by a 48-pixels gap, mimicking an erosion of the fragments. This value was the one used by Doersch et al. in \cite{doersch}. Then, we extract the center fragment and one of its 8 neighbors. Each fragment is of size $96 \times 96$ pixels, and we randomly move the location of the fragment by $\pm 7$pixels in each direction. The learning rate is 0.1 For the validation, we only consider a single pair of random fragments per image.

In the evaluation, we consider the following three setups: 1) we train the neural network on ImageNet and evaluate it on ImageNet ; 2) we train the network on ImageNet, fine-tune it on MET and evaluate it on MET (transfer setup) ; 3) we train the network from scratch on MET (MET setup). The results are compared using the accuracy of correct location prediction.

\subsection{Results}

\begin{figure}[htb]
  \centerline{\includegraphics[width=\columnwidth]{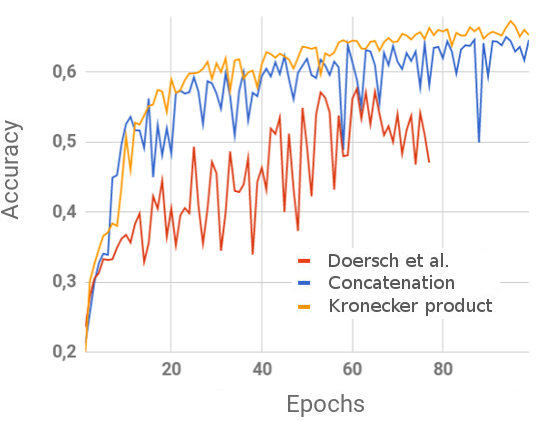}}
  \caption{Evolution of the validation accuracy on ImageNet.}
  \label{fig:acc_imagenet}
\end{figure}

On Figure \ref{fig:acc_imagenet}, we show the evolution of the validation accuracy on ImageNet for a network similar to that of \cite{doersch}, compared to our proposed architecture using either the concatenation or the Kronecker combination layer.
Our implementation of the network proposed in \cite{doersch} has fewer parameters (we reduced the fully connected layers from 4k to 512 neurons) but nonetheless outperforms what is reported in \cite{doersch}, achieving 57\% accuracy on ImageNet compared to the 40\% reported in the paper.
Our proposed architecture significantly outperforms that of \cite{doersch} by a 25\% margin.
In accordance with our intuition, the Kronecker combination consistently outperforms the concatenation combination, reaching a validation accuracy of 65\% after 100 epochs. All the results were obtained through a single run.

We show on Table \ref{tab:met} the validation accuracy of the MET dataset for both Combination Layers and comparing the MET setup to the transfer setup (training on ImageNet, validation on MET).
As we can see, training on ImageNet followed by a fine tuning on MET provides better results than training on MET alone.
In this setup, we also remark that the Kronecker combination performs significantly better than the concatenation layer, which confirms the soundness of the approach.

\begin{table}[htb]
  \centering
  \begin{tabular}{c|c|c|c|c}
     & \multicolumn{2}{c}{ImageNet $\rightarrow$ MET} & \multicolumn{2}{|c}{MET setup} \\
    & concat & kron & concat & kron \\
    \hline
    Accuracy & 59.7\% & 64.9\% & 48.9\% & 47.9\%  
  \end{tabular}
  \caption{Comparison between the transfer setup (ImageNet $\rightarrow$ MET) and the MET setup for various combination layers.}
  \label{tab:met}
\end{table}

Finally, we show on Figure \ref{fig:puzzle} examples of puzzle solving using the greedy algorithm on several images taken from the MET dataset.
As we can see, most of the predictions are correct. In the case where the algorithm wrongly predicts the position of  the fragments, we can see that the corresponding fragments are visually close to what is expected to be in that location (\textit{e.g.}, sky and could pattern in the first example).

Using this greedy algorithm, we are able to solve the jigsaw puzzle perfectly 28.8\% of the time. The proportion of correctly placed fragments is 68.8\%, which means that on average only 2 fragments are swapped per image which is consistent with the accuracy at predicting individual positions of our neural network.

\section{Conclusion}
\label{sec:ccl}

We proposed a robust deep learning method to classify the position of two neighboring fragments, which outperforms the state of the art by 25\%. We successfully apply it to solve 9-tiles puzzles, and we show promising results on a new proposed dataset composed of images taken from the Metropolitan Museum of Art.

In the future, we plan to add a ninth class describing the not-neighbor relationship. Such class will allow us to solve more challenging puzzles. We are extending our method to non-squared fragments with irregularities.

\bibliographystyle{IEEEbib}
\bibliography{refs,refs_cv}

\end{document}